\documentclass{article}
\usepackage{ijcai20}

\usepackage{times}

\usepackage{soul}
\usepackage{url}
\usepackage[hidelinks]{hyperref}
\usepackage[utf8]{inputenc}
\usepackage[small]{caption}
\usepackage{graphicx}
\usepackage{amsmath}
\usepackage{amssymb}
\usepackage{booktabs}
\usepackage{algorithm}
\usepackage{algorithmic}
\usepackage{bm}
\usepackage{color}
\usepackage[utf8]{inputenc}
\usepackage[english]{babel}

\usepackage{wrapfig}
\usepackage{multirow}
\usepackage{xcolor}
\usepackage{multicol}
\usepackage{flushend}

\renewcommand{\algorithmiccomment}[1]{\bgroup\hfill//~#1\egroup}

\title{AutoEG: Automated Experience Grafting \\
for Off-Policy Deep Reinforcement Learning}
\author{Keting Lu$^1$, Shiqi Zhang$^2$, Xiaoping Chen$^1$\\
\affiliations
$^1$ School of Computer Science, University of Science and Technology of China\\
$^2$ Department of Computer Science, SUNY Binghamton\\
\emails
ktlu@mail.ustc.edu.cn; zhangs@binghamton.edu; xpchen@ustc.edu.cn
}

\begin{document}


\maketitle

\begin{abstract}
Deep reinforcement learning (RL) algorithms frequently require prohibitive interaction experience to ensure the quality of learned policies.
The limitation is partly because the agent cannot learn much from the many low-quality trials in early learning phase, which results in low learning rate.
Focusing on addressing this limitation, this paper makes a twofold contribution. 
First, we develop an algorithm, called Experience Grafting (EG), to enable RL agents to reorganize segments of the few high-quality trajectories from the experience pool to generate many synthetic trajectories while retaining the quality.
Second, building on EG, we further develop an AutoEG agent that automatically learns to adjust the grafting-based learning strategy. 
Results collected from a set of six robotic control environments show that, in comparison to a standard deep RL algorithm (DDPG), AutoEG increases the speed of learning process by at least 30\%.
\end{abstract}

\section{Introduction}
Deep reinforcement learning (RL) algorithms recently have achieved great successes in a variety of applications, such as game playing and robot control~\cite{mnih2013playing,silver2016mastering,levine2016end}.
However, due to the high domain complexity, learning an effective action policy in such domains frequently requires a prohibitively large number of interaction samples, which significantly limits current RL algorithms' applicability.

Data augmentation is one attractive way of enabling agents to learn from insufficient data~\cite{tanner1987calculation}, and has been widely used by the machine learning community.
Augmenting data is more difficult in RL tasks, because RL agents learn from trial-and-error experience, and the ``data'' is in the form of samples of interaction experiences. 
There are at least two very different ways of augmenting interaction data for RL. 
One points to the imagination-based methods that require learning and interacting with world models to generate artificial experience, e.g.~\cite{racaniere2017imagination}. 
However, the imagination-based methods themselves are data-hungry so as to ensure the quality of the learned world models. 
Hindsight experience replay (HER)~\cite{andrychowicz2017hindsight} is another way of augmenting data for RL agents, where the agents synthesize samples \emph{without} computing world models. 
However, the HER method is only applicable to domains where the goal condition is explicitly defined in the state space.
For instance, HER was originally applied to a domain of a robot arm moving from one point to another, where the goal corresponds to a position in the 2D state space.
In line with the HER method, we aim at post-processing experiences to generate ``successful'' samples. 
Beyond that, we focus on more challenging domains, where \emph{goal states} do not exist. 
For instance, the \emph{Walker2D} task in MuJoCo~\cite{todorov2012mujoco} requires the agent running as fast as possible, rendering the original HER method inapplicable.

In this paper, we develop an algorithm, called {\bf Experience Grafting (EG)}, for generating high-quality, synthetic trajectories to speed up the agent's learning process.
In comparison to HER that manipulates individual trajectories, EG searches for pairs of trajectory segments, where one's ``head'' state and the other's ``tail'' state are of sufficient similarity.
Moreover, we develop an {\bf Automated EG (AutoEG)} algorithm that enables the RL agent to learn to dynamically adjust its grafting strategy.
For instance, AutoEG enables an experienced agent to be very ``picky'' in grafting, because it is able to produce good-quality samples from its own interaction experience.

\begin{figure*}[t]
  \begin{center}
  \hspace*{-.5em}
  \includegraphics[width=0.9\textwidth]{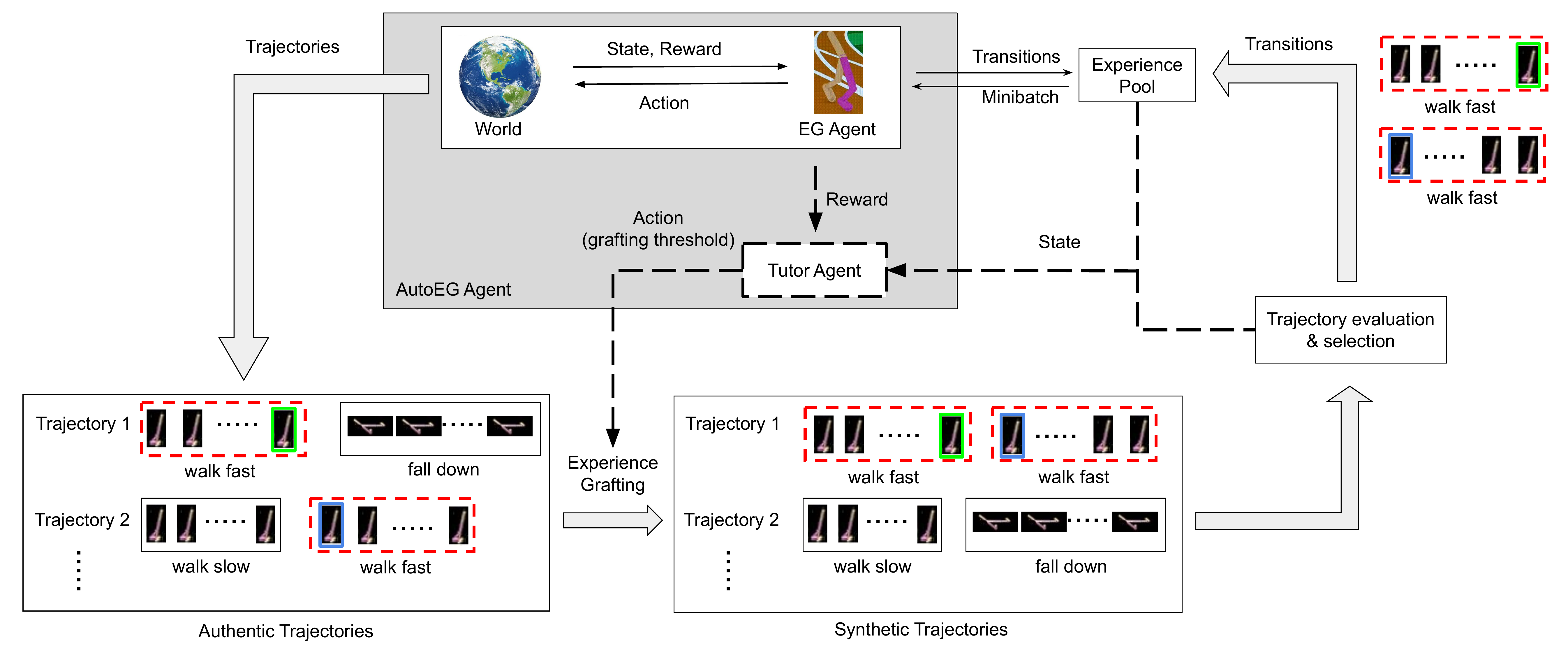}
  \vspace{-.5em}
  \caption{Experience Grafting (EG) enables RL agents to utilize authentic trajectories of mixed qualities to generate many synthetic trajectories, and selectively add the high-quality ones into the experience pool for learning action policies. 
  Automated EG (AutoEG), which combines EG and ``Tutor Agent'',  enables the learning agent to adjust its grafting strategy throughout the learning process. }
  \label{fig:framework}
  \end{center}
  \vspace{-1em}
\end{figure*}

EG and AutoEG are generally applicable to off-policy RL algorithms, such as the value-based~\cite{mnih2013playing} and policy gradient methods~\cite{sutton2000policy,schulman2017proximal} among others.
We use deep deterministic policy gradient (DDPG)~\cite{lillicrap2015continuous}, which well accounts for continuous action spaces, for both learning the policy of interacting with the environments and learning the grafting policy, as shown in Figure~\ref{fig:framework}.
We have evaluated EG and AutoEG using six Roboschool environments, an extension to MuJoCo~\cite{todorov2012mujoco}.
We used a metric called \emph{area under the learning curve} (AUC) to evaluate the speed of the learning process~\cite{taylor2009transfer,stadie2015incentivizing}. 
Results suggest that: 1)~Although EG performs better than standard DDPG in most environments, its performance is sensitive to its handcrafted grafting strategy; and 2)~AutoEG further enables the capability of grafting policy learning, and produces better performance (c.f., standard DDPG) in learning speed by at least 30\% in all six Roboschool environments.



\section{Related Work}
Data augmentation is widely used for enlarging training sets, and has become an essential step of mitigating over-fitting in supervised learning~\cite{krizhevsky2012imagenet,zhang2015character,jaitly2013vocal}.
For instance, in image classification, linear transformations are commonly applied~\cite{eigen2014depth}, whereas text classification researchers often rephrase words or phrases with their synonyms~\cite{zhang2015character}.
There is the recent effort on automated architecture search for data augmentation, where an AutoML approach was used to automatically generate data augmentation policies~\cite{cubuk2018autoaugment}.
In comparison to these methods for supervised learning, this work is on augmenting trial-and-error experiences for RL agents.

Researchers have developed RL and planning methods to learn world models, and simulate experiences of agent interacting with the real world. 
One notable example of these methods is Dyna-Q~\cite{sutton1990integrated}.
The agent learns an approximation function (e.g., neural networks) from experiences, and uses the (imperfect) world model to generate artificial trajectories to augment experiences~\cite{racaniere2017imagination,buckman2018sample}.
Other examples include~\cite{pascanu2017learning,peng2018deep,su2018discriminative,kalweit2017uncertainty}, as well as the imagination-based methods, e.g.,~\cite{racaniere2017imagination}.
These approaches avoid the need of extensive interaction experience with the real world.
However, these methods tend to be sensitive to the learned model's quality, because an inaccurate model has a detrimental effect on performance~\cite{gu2016continuous}.
Our EG and AutoEG algorithms provide an alternative way of generating artificial experiences for better learning performances, while avoiding the challenging task of learning world models.

Hindsight experience replay (HER) enables an agent to augment trajectories by replaying each episode with a goal that is different from the goal that the agent was originally trying to achieve~\cite{andrychowicz2017hindsight}.
HER provides a new way of augmenting interaction experiences for RL agents.
More recent work has applied HER to dialogue domains, enabling the manipulation of trajectory segments to form new dialogues~\cite{dialogue}. 
However, the HER-style methods has the limited applicability, requiring the goal condition being defined in the state space (e.g., the box's target location in ``pushing'' task, and slots being filled in goal-oriented dialog agents). 
In comparison, our developed EG and AutoEG methods support tasks where goal states do not exist, e.g., the Walk2D task where the agent tries to walk fast, rendering the HER methods inapplicable.

Automated machine learning (AutoML) recently has emerged as a new area in the machine learning community~\cite{quanming2018taking}.
AutoML has been successfully applied to problems including neural architecture search (NAS)~\cite{zoph2017neural} and hyper-parameter optimization~\cite{feurer-automlbook18a}.
Inspired by the AutoML idea, we formulate the hyper-parameter search of EG as a separate RL problem, enabling one RL agent to learn a policy to tune the (experience grafting) parameter of another RL agent.

\section{Algorithm}
In this section, we describe our automated experience grafting (AutoEG) algorithms.
Figure~\ref{fig:framework} summarizes the two learning agents within AutoEG: 
the EG agent learns from the trial-and-error experience with the environment, as well as the artificial trajectories from experienc grafting; and the Tutor agent learns to adjust the EG agent's grafting strategy on the fly. 
For instance, Tutor tends to make an ``experienced'' EG agent more cautious in utilizing synthetic trajectories, because the EG agent itself is able to produce high-quality trajectories. 

\subsection{Functions for Experience Grafting}
\label{sec:func}
We use the term of \emph{trajectory segment} (or simply \emph{segment}) to refer to a sequence of \emph{transitions}, where each segment includes at least one transition.
Each transition, $tr$, is in the form of a state-action-reward-state tuple, i.e., $tr=(s,a,r,s\_)$, where $s$ (or $s\_$) is the current (or next) state, $r$ is the reward, and $a$ is the action.
A \emph{trajectory} is a segment whose first transition starts with the initial state, and whose last transition leads to an terminal state.

\paragraph{Distance Function:}
We introduce a distance function to measure the similarity between two states:
$$
  Dis(s,s\_) = W\big(\mathbb{P}(s), \mathbb{P}(s\_)\big),
$$
where function $\mathbb{P}$ normalizes a state vector to a distributional representation, and $W(\mathbb{P}_1,\mathbb{P}_2)$ is the \emph{1st Wasserstein distance} or \emph{earth mover's distance}~\cite{rubner1998metric}.
$$
\displaystyle W(\mathbb{P}_1, \mathbb{P}_2)= \inf_{\gamma\in \Pi (\mathbb{P}_1, \mathbb{P}_2 )} \mathbf {E}_{(x,y)\sim \gamma} {\big [} \| x-y \| {\big ]},
$$
where $\Pi(\mathbb{P}_1,\mathbb{P}_2)$ is the set of all joint distributions $\gamma(x,y)$ whose marginals are respectively $\mathbb{P}_1$ and $\mathbb{P}_2$.
Intuitively, $\gamma(x,y)$ indicates how much ``mass'' must be transported from $x$ to $y$ in order to transform the distributions $\mathbb{P}_1$ into distribution $\mathbb{P}_2$.
The earth mover's distance then is the ``cost'' of the optimal transport plan.


\paragraph{Error Function:}
We say a segment is a \emph{head} (\emph{tail}) segment, if the last (first) transition of this segment is used for grafting.
Accordingly, we graft head segments to tail segments.
Given head segment $Seg_{1}$ and tail segment $Seg_{2}$, the grafting error is defined as below
$$
  Err(Seg_1, Seg_2) = Dis\big(Term(Seg_1), Init(Seg_2)\big),
$$
where $Term(Seg)$ returns the terminal state of $Seg$, and $Init(Seg)$ returns the initial state of $Seg$.
A smaller grafting error indicates that the generated synthetic trajectory is more realistic.
Synthetic trajectories of high grafting errors are of low \emph{grafting quality}.


\paragraph{Union Function:}
Given two segments $Seg_1$ (head) and $Seg_2$ (tail), the grafting union function generates a synthetic trajectory that grafts $Seg_1$ to $Seg_2$, if their grafting error is lower than $\epsilon$, the grafting threshold. 
It should be noted that $\epsilon$ plays an important role in later sections. 
\begin{align*}
    &Uni(Seg_1, Seg_2) = \\
    &\begin{cases} append(Seg_1,Seg_2)~~~~~ & Err(Seg_1,Seg_2) < \epsilon \\
                     \emptyset & otherwise
       \end{cases}
\end{align*}
The grafting union function ($Uni$) can be used for generating potentially many synthetic trajectories, and we selectively use only the ``high-quality'' trajectories for learning purposes.

\paragraph{Performance Quality Function:}
We use the accumulative reward of a trajectory to measure the trajectory's \emph{performance quality}. 
$$
  Qua(Trj) = R_0(Trj)
$$

It should be noted that a trajectory's quality can be evaluated in two ways, i.e., being realistic and being effective to learning.
For instance, a trajectory of a walker instantly changing its position from lying on the floor to jumping in the air is unrealistic, and hence is of poor \emph{grafting quality}.
A trajectory of a walker lying on the floor without any movement has good grafting quality, but its \emph{performance quality} is poor, because an RL agent can hardly learn much from it.

\paragraph{Grafting Function:}
Given an authentic trajectory, $auTrj$, and a set of synthetic trajectories, $\textbf{syTrj}'$, we use $\textbf{syTrj}$ to represent the set of synthetic trajectories whose quality is higher than that of the authentic trajectory.
To avoid the grafting function generating too many trajectories, we introduce $\Theta$, the maximum number of synthetic trajectories allowed given one authentic trajectory.
In case more than $\Theta$ trajectories are qualified, the grafting function $G$ sorts the trajectories using their qualities, and outputs the top $\Theta$ trajectories:
\begin{align*}
&\textbf{syTrj} = \\
&\{syTrj~|~Qua(syTrj) \geq Qua(auTrj), ~syTrj \in \textbf{syTrj}'\},\\
    & G(auTrj, \textbf{syTrj})=
        \begin{cases}
            \textbf{syTrj}[1:\Theta]   &\textnormal{if}~~ |\textbf{syTrj}| > \Theta\\
            \textbf{syTrj} & otherwise
        \end{cases}
\end{align*}
While using this grafting function, we force the set of synthetic trajectories to be those that are generated by grafting with the authentic trajectory.
The union and grafting functions together ensure that the synthetic trajectories used for RL are both realistic and potentially effective for learning.


\vspace{1.5em}
\begin{algorithm}[t]
\scriptsize
\caption{Experience Grafting}
\label{alg:EG}
\begin{algorithmic}[1]

\REQUIRE
$\epsilon$ (grafting threshold), $T$ (authentic trajectory), $Lib$ (segment library), $N^{ext}$ (number of extracting positions), and $N^{gft}$ (number of grafting positions)
\ENSURE
a set of synthetic trajectories
\FOR[Start: Segment extraction]{$i \gets 1$ to $N^{ext}$} 
\STATE Randomly select position $p=random(0,size(T))$
\STATE Extract transition $T[p]=(s_p, a_p, r_p, s_{p+1})$ from $T$
\STATE Add $s_p: T[p:size(T)]$ into $Lib$ as a new indexed segment, where $s_p$ is the key and $T[p:size(T)]$ is the value
\label{line:lib}
\ENDFOR \COMMENT{End: Segment extraction}
\FOR[Start: Trajectory synthesis]{$i \gets 1$ to $N^{gft}$}
\STATE Randomly select position $q=random(0,size(T))$
\STATE Extract transition $T[q]=(s_q, a_q, r_q, s_{q+1})$
\STATE Search $Lib$ for segments using $s_{q+1}$: $\textbf{Seg} = Lib.\textnormal{get}(s_{q+1},\epsilon)$
\STATE Initialize a empty set of synthetic trajectory $\mathbf{syTrj}=\emptyset$
\FOR{$seg \in \mathbf{Seg}$}
\STATE synthetic trajectory $syTrj= Uni\big(T[0:q], seg\big)$
\label{line:syn}
\STATE $\mathbf{syTrj} \leftarrow \mathbf{syTrj} ~\cup~ syTrj $
\ENDFOR
\ENDFOR \COMMENT{End: Trajectory synthesis}
\STATE return $G(T, \textbf{syTrj})$, where $G$ is the grafting function

\end{algorithmic}
\end{algorithm}
\vspace{-1.5em}

\subsection{Experience Grafting}
Algorithm~\ref{alg:EG} presents our experience grafting (EG) algorithm that includes two phases for \emph{segment extraction} and \emph{trajectory synthesis} respectively.
The input of EG includes $\epsilon$ (grafting threshold), $T$ (an authentic trajectory), $Lib$ (a segment library), and two parameters of $N^{ext}$ and $N^{gft}$.
%

Lines 1-5 in Algorithm~\ref{alg:EG} presents the steps for segment extraction.
There are $N^{ext}$ iterations in this phase, where one segment is generated in each iteration.
Position $p$ is randomly selected in trajectory $T$.
From the $p$ position, $T$ is cut into two segments, where EG saves the tail (i.e., from position $p$ to the end) to segment library $Lib$.
EG uses the initial state $s_p$ as the key for indexing, because segments in $Lib$ will be used for grafting as the tail segment in later steps.
It should be noted that EG only presents the steps of processing one authentic trajectory.
In practice, EG is repeatedly called whenever a new authentic trajectory comes in.
As a result, in most cases, $Lib$ already includes many segments each time EG is activated.



The segment library, $Lib$, stores segments and indexes the segments using their initial state, where EG discretizes the state space for the indexing purpose.
This operation ensures efficient search in $Lib$, and is important from the practical perspective.



Lines 6-15 in Algorithm~\ref{alg:EG} presents the trajectory synthesis steps.
EG randomly selects a transition in trajectory $T$ in Line 7, and uses $s_{q+1}$, the resulting state of $T$, to search for segment candidates for grafting in $Lib$ (Line 9).
Entering the inner for-loop (Lines 11-14), in each iteration, EG uses the union function ($Uni$) to generate one or zero synthetic trajectory ($syTrj$) with the guarantee that the generated synthetic trajectory is realistic.
$\textbf{syTrj}$ saves a set of realistic synthetic trajectories.
Finally, EG uses the grafting function to output a set of synthetic trajectories that are of both good \emph{performance quality} and good \emph{grafting quality} (defined in Section~\ref{sec:func}).
%

\textit{Remark:}
EG enables an RL agent to more efficiently utilize its trial-and-error experiences by synthesizing good-quality trajectories.
$\epsilon$ is an important parameter in this grafting process that directly determines how many trajectories can be synthesized, as well as how realistic these trajectories are.
Intuitively, when $\epsilon$ is small (e.g., close to zero), the generated trajectories are very realistic (i.e., can hardly be distinguished from the authentic ones), but the issue of small $\epsilon$ values is that very few trajectories can be synthesized.
When $\epsilon$ is large, more synthetic trajectories can be generated, but they might look very artificial, e.g., a robot lying on the floor instantly changes its position 
to be jumping in the air, which is certainly detrimental to agent learning.
The trade-off between synthetic trajectories' quality and quantity motivates the development of AutoEG for learning to adjust $\epsilon$ for adaptive learning behaviors.

\begin{figure*}[t]
  \begin{center} 
  \includegraphics[width=.9\textwidth]{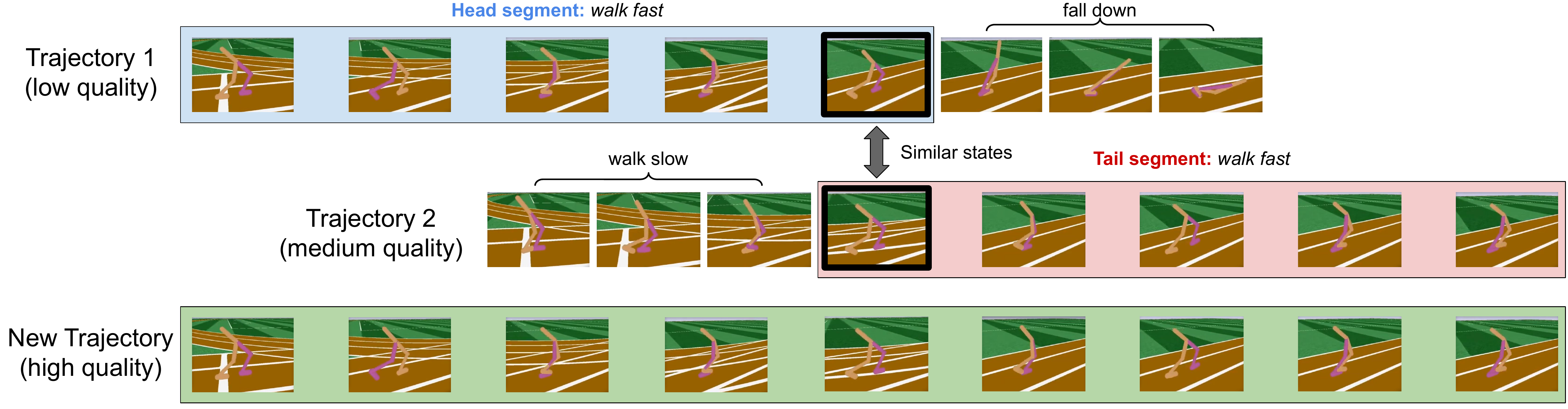}
  \caption{An illustrative example of the experience grafting process in the \texttt{\small Walker2d} environment. Trial-1 is of low quality because of the walker's fall-down behavior. Trial-2 is of medium quality due to its slow-walking behavior at the beginning. EG located the two trials, identified the ``grafting'' position, and generated a high-quality trial, where the walker maintains a high speed throughout the trial. }
  \label{fig:example}
  \end{center}
  \vspace{-.8em}
\end{figure*}

\begin{algorithm}[t]
\scriptsize
\caption{Automated EG (AutoEG)}
\label{alg:FWEG}

\begin{algorithmic}[1]
\REQUIRE
$M$ (number of episodes), $N^{EG}$ (size of minibatch of the EG agent), $N^{Tutor}$ (size of minibatch of the Tutor agent), 
Maximal number of Synthetic Trajectory in one time $\Theta$, Horizon $H$
\STATE Initialize two DDPGs for the EG and Tutor agents, including their replay buffers of $R^{EG}$, and $R^{Tutor}$
\STATE Initialize $horizon=0$, Tutor's current state, $s^{Tutor}=(0,0)$; and EG's sum of rewards in horizon H, $Sum\_of\_Reward=0$
\FOR{$episode \gets 1$ to $M$}
\STATE Initialize trajectory $\mathcal{T}=\emptyset$, and receive initial observation state $s_1$
\COMMENT{Start: the EG agent}
\WHILE{current state $s_t$ is not terminal}
\STATE Select action $a_t=\pi^{EG}(s_t)$ for the EG agent, and execute $a_t$, and observe reward $r_t$ and new state $s_{t+1}$
\STATE Store transition $(s_t, a_t, r_t, s_{t+1})$ in $R^{EG}$
\STATE $Sum\_of\_Reward \leftarrow Sum\_of\_Reward + r_t$, and records current transition $\mathcal{T}=\mathcal{T} \cup (s_t, a_t, r_t, s_{t+1})$
\STATE Sample a random minibatch of $N^{EG}$ transitions $(s_i, a_i, r_i, s_{i+1})$ from $R^{EG}$ and update $\pi^{EG}$, the policy of the EG agent
\ENDWHILE 
\STATE Get current grafting threshold using the Tutor agent's action policy, $\epsilon=\pi^{Tutor}(s^{Tutor})$
\STATE Call Algorithm~\ref{alg:EG} to generate a set of synthetic trajectories, $\mathbf{syTrj}$, using $\epsilon$
\FOR{$syTrj \in \mathbf{syTrj}$}
\STATE Store transition $(s_t, a_t, r_t, s_{t+1})$ in $syTrj$ into $R^{EG}$
\STATE Update the EG agent's policy $\pi^{EG}$ with a minibatch of $N^{EG}$ transitions from $R^{EG}$
\ENDFOR \COMMENT{End: the EG agent}
\STATE Update $r^{trs}$, the ratio of transitions from synthetic trajectories to all transitions in $R^{EG}$ ~~~~~~\,\,\,\,\,\,// Start: the Tutor~agent
\STATE Update $r^{trj}$, the ratio of generated synthetic trajectories to $\Theta$, i.e., $r^{trj} = len(\mathbf{syTrj}) / \Theta$
\STATE Get the Tutor agent a new state, $s^{Tutor}_{\_} \leftarrow (r^{trs},r^{trj})$
\IF{ $horizon < H$ }
\STATE Store transition $(s^{Tutor}, \epsilon, 0, s^{Tutor}_{\_} )$ in $R^{Tutor}$, and update $s^{Tutor} \leftarrow s^{Tutor}_{\_}$ and $horizon \leftarrow horizon + 1$
\ELSE 
\STATE Store the following transition in $R^{Tutor}$
$$(s^{Tutor}, \epsilon, \frac{Sum\_of\_Reward}{H}, s^{Tutor}_{\_})$$
\STATE Remove all synthetic transitions from $R^{EG}$; and initialize $s^{Tutor}=(0,0)$; $horizon \leftarrow 0$; and $Sum\_of\_Reward \leftarrow 0$
\ENDIF 
\STATE Update $\pi^{Tutor}$, the policy of Tutor, with a minibatch of $N^{Tutor}$ transitions from $R^{Tutor}$
\COMMENT{End: the Tutor~agent}
\ENDFOR
\end{algorithmic}
\end{algorithm} 

\subsection{AutoEG: Learning Grafting Strategies}

Recent research on AutoML~\cite{quanming2018taking} and neural architecture search~\cite{zoph2017neural} has shown promising results on ``learning to learn'' methods.
In line with these methods, we develop another learning agent, called \emph{Tutor}, that guides the EG agent by learning to adjust its grafting strategy throughout the EG agent's trial-and-error process.
We use Automated EG (AutoEG) to refer to the combination of Tutor and EG.
The Tutor agent is modeled as a DDPG-based RL problem, which is defined below.

\paragraph{State:}
The Tutor agent only exchanges information with the EG agent, and has no direct interaction with the working environment.
Tutor's state space is specified using two state features: 
\vspace{-.2em}
\begin{itemize} 
    \item The ratio of transitions from synthetic trajectories to all transitions in the replay buffer; and \vspace{-.2em}
    \item The ratio of synthetic trajectories from the grafting function to $\Theta$, the maximum number of synthetic trajectories allowed given one authentic trajectory (Sec.~\ref{sec:func}). \vspace{-.2em}
\end{itemize}

\paragraph{Action:} The Tutor agent's action is in the form of a real number, ranging from $0.0$ to $1.0$, where an action directly corresponds to a grafting threshold ($\epsilon$).

\paragraph{Reward:} We define a finite time horizon for the interactions between the Tutor agent and the EG agent.
Tutor's reward function is defined as the average of the rewards the EG agent receives within this horizon.
Accordingly, the goal of this Tutor agent is to maximize the accumulative reward of the EG agent in its next $H$ steps, where $H$ is the time horizon. 
To be more specific, each step of the Tutor agent corresponds a complete episode of the EG agent interacting with the environment.

It makes sense to model the Tutor agent as an RL problem, because the tutor agent's reward is noisy and is not immediate.
As a result, it is necessary for the Tutor agent to learn an action policy to tune the EG agent's grafting threshold in a way that the EG agent's long-term reward is maximized.

Algorithm~\ref{alg:FWEG} presents our Automated EG (AutoEG) framework. 
There are two agents being learned, namely the EG agent and the Tutor agent. 
The EG agent is responsible for interacting with the environment for a certain task (Lines 4-16), while Tutor learns a grafting policy for the EG agent (Lines 17-26). 
Synthetic trajectories are generated once each authentic trajectory is sampled, and are used for updating EG's policy by augmenting its replay buffer (Lines 11-16). 
Tutor's each episode corresponds to the EG agent interacting with the environment for $H$ times, i.e., EG's $H$ episodes. 
After the EG agent completing each of its $H$'th episodes, all the synthetic trajectories augmented in its replay buffer are cleared to get prepared for the Tutor agent's next episode. 


\section{Experiment}
\label{sec:exp}
EG and AutoEG have been evaulated using six robotic environments in Roboschool from OpenAI.\footnote{{\small \texttt{https://github.com/openai/roboschool}}}
In this section, we introduce the experimental setup, and implementation details of EG and AutoEG, where both EG and Tutor agents are implemented using DDPG as the RL algorithm.
Although AutoEG is generally applicable to off-policy RL algorithms, we selected DDPG in the evaluation due to its relatively longer history and (arguably) better popularity. 
Our main evaluation metric is called \emph{area under the learning curve} (AUC), which has been used for evaluating the speed of the learning process~\cite{taylor2009transfer,stadie2015incentivizing}. 
Another metric used in this work is called \emph{policy quality}, which is measured by taking the average of total rewards over the last $1000$ episodes. 

Our main hypotheses include:
I) Well-tuned EG agents perform better than a standard DDPG agent (referred to as no-EG) in AUC, but their performances can be sensitive to the handcrafted grafting strategies; and
II) AutoEG produces the best performance in both learning speed and policy quality in comparison to EG and no-EG methods.



\begin{figure*}[t]
  \begin{center} 
  \vspace{-3em}
  \hspace*{-3em}
  \includegraphics[width=1.1\textwidth]{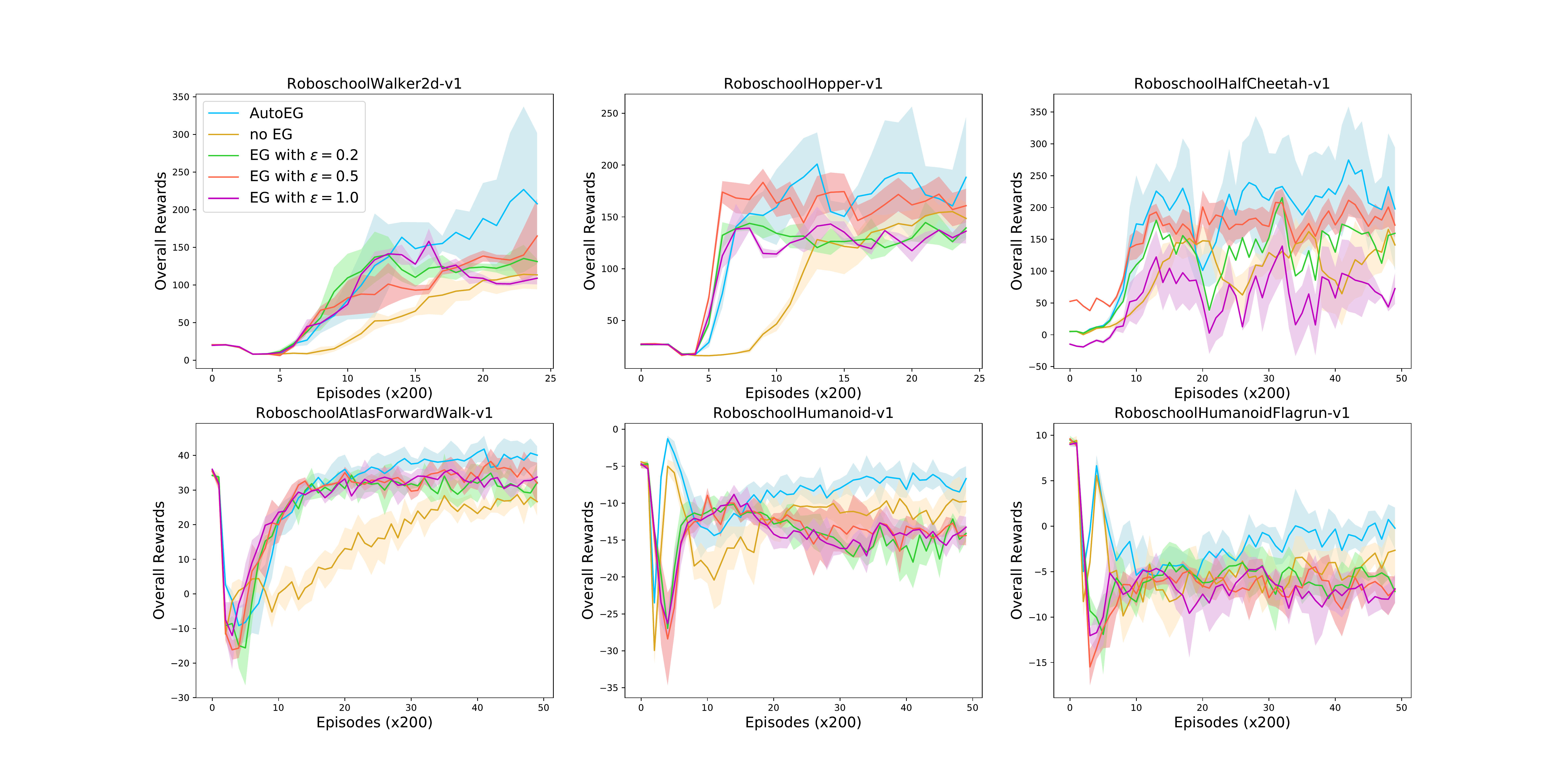}
  \vspace{-3em}
  \caption{Overall rewards of AutoEG, EG with different grafting thresholds, and a no-EG baseline (vanilla DDPG). 
  EG agents perform better than ``no EG'' in most cases, and AutoEG performs the best in all environments. 
  Table~\ref{tab_results} presents the quantitative analysis. }
  \label{e1}
  \end{center}
\end{figure*}

\begin{table*}[h]  
\scriptsize
  \centering 
  \caption{{\bf AutoEG improves the performance in AUC by at least 30\% in comparison to the no-EG baseline (DDPG)}. AutoEG performs the best in policy quality, while EG's performance is sensitive to the $\epsilon$ value. }  
  \vspace{-.8em}
  \label{tab_results}  
    \begin{tabular}{|c||c|c|c|c||c|c|c|c|c|}  
    \hline  
    \multirow{2}{*}{Environments}&  
    \multicolumn{4}{c||}{AUC Improvement (\%)} & \multicolumn{5}{c|}{Policy Quality}\cr\cline{2-10}  
    &$\epsilon=.2$ (EG) &$\epsilon=.5$ (EG) &$\epsilon=1.0$ (EG) &AutoEG &no EG    &$\epsilon=.2$ (EG) &$\epsilon=.5$ (EG) &$\epsilon=1.0$ (EG)  &AutoEG\cr  
    \hline  
    \hline  
    Wakler2d&62.6&51.3&53.4&{\bf 99.5}     &110.3&127.9&142.3&105.2&{\bf 202.5}\cr\hline
    Hopper&27.2&58.6&24.9&{\bf 57.4}       &150.1&135.3&163.3&129.9&{\bf 176.0}\cr\hline
    HalfCheetah&22.9&65.4&-39.7&{\bf 84.2}      &139.0&145.7&181.8&65.0&{\bf 207.1}\cr\hline
    AtlasForwardWalk&66.8&78.7&78.3&{\bf 92.5}     &27.2&30.7&34.5&32.3&{\bf 39.6}\cr\hline
    Humanoid&-12.7&-10.0&-13.2&{\bf 30.7}        &-10.2&-14.5&-13.7&-14.4&{\bf -7.5}\cr\hline
    HumanoidFlagrun&-16.2&-36.2&-37.9&{\bf 64.5}       &-3.3&-5.8&-6.7&-7.8&{\bf -0.1}\cr\hline
    
    \end{tabular}  
    \vspace{-1em}
\end{table*}


\subsection{Implementation Details}
Six robotic environments from Roboschool have been used for evaluations.
Relatively low-dimensional environments include \texttt{\small Walker2d-v1}, 
\texttt{\small HalfCheetah-v1}, 
\texttt{\small Hopper-v1}, 
while the others are high-dimensional, more challenging environments, including \texttt{\small AtlasForwardWalk-v1}, \texttt{\small Humanoid-v1}, and \texttt{\small HumanoidFlagrun-v1}.

In all tasks, we follow the original DDPG architecture (for both EG and Tutor agents), including the hyper-parameters and network initialization~\cite{lillicrap2015continuous}.
In the EG implementations, the replay buffer size is $1e6$, and experience replay starts when the buffer size reaches $1e4$.
In the Tutor implementation, the replay buffer size is the same, while experience replay starts when the size of buffer reaches $100$.
The experience reply starts late in EG, because it cannot adjust its grafting strategy and replaying a small amount of experience is detrimental.
The size of minibatch is 16 in the DDPG-EG setting, and  it is 10 in the DDPG-AutoEG setting (time horizon $H$ is 10). The value of $\Theta$ is 5.
There is the discretization in $Lib$ (segment library) to allow efficient search, where each bin is sized $1.0$ in each dimension, and includes at most $1000$ segments.
Each data point in the figures of this paper is an average of five runs, where we also report the standard deviations.

\subsection{Illustrative Example}
\label{sec:example}

Figure~\ref{fig:example} shows an example of experience grafting process in the \texttt{\small Walker2d} environment. 
The segment library $Lib$ saves a set of trajectory segments (Line~\ref{line:lib} in Algorithm~\ref{alg:EG}), where each segment is indexed using its initial state. 
In this case, the very left state of the ``pink'' (second half) segment in Trajectory 2 was used a key state for indexing. 
Given the key state and its ``Tail segment'', EG searched over all new trajectories, and found Trajectory 1 has a state of significant similarity, while the corresponding ``Head segment'' is of good quality. 
EG then initiated the grafting (Line~\ref{line:syn} in Algorithm~\ref{alg:EG}), and produced the synthetic trajectory of high quality. 


\subsection{Experimental Results}
Figure~\ref{e1} shows the experimental results from comparisons among no-EG, EG (with different $\epsilon$ values), and AutoEG agents.
From all the curves, we see the EG agents perform better than naive DDPG (corresponding to the ``no EG'' curves) in most cases, and AutoEG performs the best in all environments. 
A side observation is that the same grafting threshold produces different performances in different environments.
For example, in \texttt{\small Walker2d}, the grafting threshold of $0.5$ produces the best performance whereas in \texttt{\small HumanoidFlagrun}, $0.2$ works better.
In addition, the same grafting threshold might be good in early learning phase, but is not as good when the agent has more online experiences.
For instance, in the \texttt{\small Walker2d} environment, $1.0$ is initially a good grafting threshold, but not anymore after $3k$ episodes.
This observation further justifies the need of the Tutor agent that learns to tune the EG agent's grafting threshold at runtime.
For no-EG, EG, and AutoEG agents, the ``Overall Rewards'' (y-axis of Figure~\ref{e1}) reflects the accumulative reward from interactions within an episode.

Tables~\ref{tab_results} is used to evaluate Hypotheses I and II for quantitative results. 
Table~\ref{tab_results} shows the quantitative improvements on both training speed and final performance, where we use the metrics of AUC\footnote{Formally, the AUC improvement is computed using $(A-B)/|B|$, where $A$ and $B$ are areas under the two curves respectively, and $B$ is the baseline method's area. } and policy quality in evaluations, as defined in the beginning of this section. 
The ``AUC Improvement \%'' is computed by comparing EG and AutoEG agents' performances with the no-EG baseline.
For instance, in \texttt{\small Walker2d}, the AUC improvement produced by AutoEG is almost $100\%$, which means AutoEG almost doubled the accumulative rewards over the same number of episodes in comparison to a standard DDPG approach. 
From the left of Table~\ref{tab_results}, we see EG agents with different grafting thresholds perform better than the no-EG baseline in most environments (Hypothesis I), and AutoEG produces at least 30\% improvements in AUC (up to 99.5\% in Walker2d) in comparison to the no-EG baseline (Hypothesis II). 

The right of Table~\ref{tab_results} shows that AutoEG also dominates the performance in the quality of the ultimately learned policies in all six environments. 
In some environments, the EG agents' performance in policy quality is very sensitive to the selection of the grafting threshold. 
For instance, in \texttt{\small HalfCheetah}, EGs with different grafting thresholds produced policy qualities ranging from 65.0 to 145.7, which further justifies the need of the Tutor agent for learning the grafting strategy.

\section{Conclusions and Future Work}
We develop an experience grafting (EG) algorithm that enables off-policy reinforcement learning (RL) agents to synthesize and learn from many good-quality trajectories, which identifies the first contribution of this paper.
The second contribution is on automated EG (AutoEG), where we develop another RL agent (Tutor agent) that learns to adjust the grafting strategy from the EG agent's learning experience. 
Experiments were conducted using the DDPG architecture in Roboschool environments.
EGs performed better than no-EG baselines in both learning rate and policy quality in most cases, and AutoEG performs the best in all environments. 

AutoEG only allows two-segment grafting. 
In the future, we would like to study the feasibility of experience grafting with more than two trajectory segments. 
We expect better performance in learning rate with multi-segment EG, though there will be the scalability challenge that must be addressed. 


{\small
\bibliographystyle{named}  
\bibliography{references}  
}

\end{document}